\documentclass{article} % For LaTeX2e
\usepackage[english]{babel}
\usepackage[accepted,nohyperref]{icml2019} 
\usepackage{times}
\usepackage{hyperref}
\usepackage{amsmath,amsfonts,amssymb,amsthm}
\usepackage{url}
\usepackage{graphicx}

\usepackage{algorithm}
\usepackage{algorithmic}
\usepackage{listings}
\usepackage{multirow}
\usepackage{tabularx}

\usepackage{xcolor}         % hyperlinks
\definecolor{LinkColor}{rgb}{0.0, 0.18, 0.39}
\hypersetup{colorlinks=true,
    linkcolor=LinkColor,
    citecolor=LinkColor,
    filecolor=LinkColor,
    menucolor=LinkColor,
    urlcolor=LinkColor}

\title{Contextual Salience \\for Fast and Accurate Sentence Vectors}

\begin{document}

\twocolumn[
\icmltitle{Contextual Salience for Fast and Accurate Sentence Vectors}

\begin{icmlauthorlist}
\icmlauthor{Eric Zelikman}{st}
\icmlauthor {Richard Socher}{sf}
\\
\texttt{ezelikman@cs.stanford.edu}, \texttt{richard@socher.org}
\end{icmlauthorlist}

\icmlaffiliation{st}{Stanford University}
\icmlaffiliation{sf}{Salesforce Research}

\icmlcorrespondingauthor{Eric Zelikman}{ezelikman@cs.stanford.edu}

\icmlkeywords{NLP}
\vskip 0.3in
]

\printAffiliationsAndNotice{} 

\begin{abstract}
Unsupervised vector representations of sentences or documents are a major building block for many language tasks such as sentiment classification.
However, current methods are uninterpretable and slow or require large training datasets. 
Recent word vector-based proposals implicitly assume that distances in a word embedding space are equally important, regardless of context.
We introduce contextual salience (CoSal), a measure of word importance that uses the distribution of context vectors to normalize distances and weights.
CoSal relies on the insight that unusual word vectors disproportionately affect phrase vectors. 
A bag-of-words model with CoSal-based weights produces accurate unsupervised sentence or document representations for classification, requiring little computation to evaluate and only a single covariance calculation to ``train." 
CoSal supports small contexts, out-of context words and outperforms SkipThought on most benchmarks, beats tf-idf on all benchmarks, and is competitive with the unsupervised state-of-the-art.
% As a result of its computational polynomial dependence on vocabulary size, it makes real-time learning with high-quality context-aware sentence embeddings feasible.  
% Further, we present alternative applications of CoSal including a measure of word similarity in context. 
% In addition, we will release code for training and testing CoSal and its use in classification.
\end{abstract}

\section{Introduction}
%TODO: try not to have long sentence. short and sweet and clear!
Global context, a representation of an analyzed text's nature, such as recognizing that one is reading a movie review, is useful for transfer learning and document interpretation \cite{AAAI148361, 2018arXiv180205365P}. Yet, current solutions that account for global context are black boxes: an algorithm takes a document and returns some function mapping word vectors to a context representations. Method outputs vary from a vector \cite{2017arXiv170800107M} to another deep structure \cite{2018arXiv180205365P}. 
%TODO is that what you mean correct?
%Yes, exactly 
This results in obfuscated approaches to using context that are uninterpretable and resistant to further extension. Further, these algorithms often still require thousands of sentences in a transfer dataset to train effectively, especially when unsupervised \cite{2017arXiv170800107M}.  
We introduce CoSal, an approach relying on normalization of distances in a semantic space with respect to the word vector distribution of a context.

\begin{figure}[t!]
\begin{center}
\includegraphics[width=0.5 \textwidth]{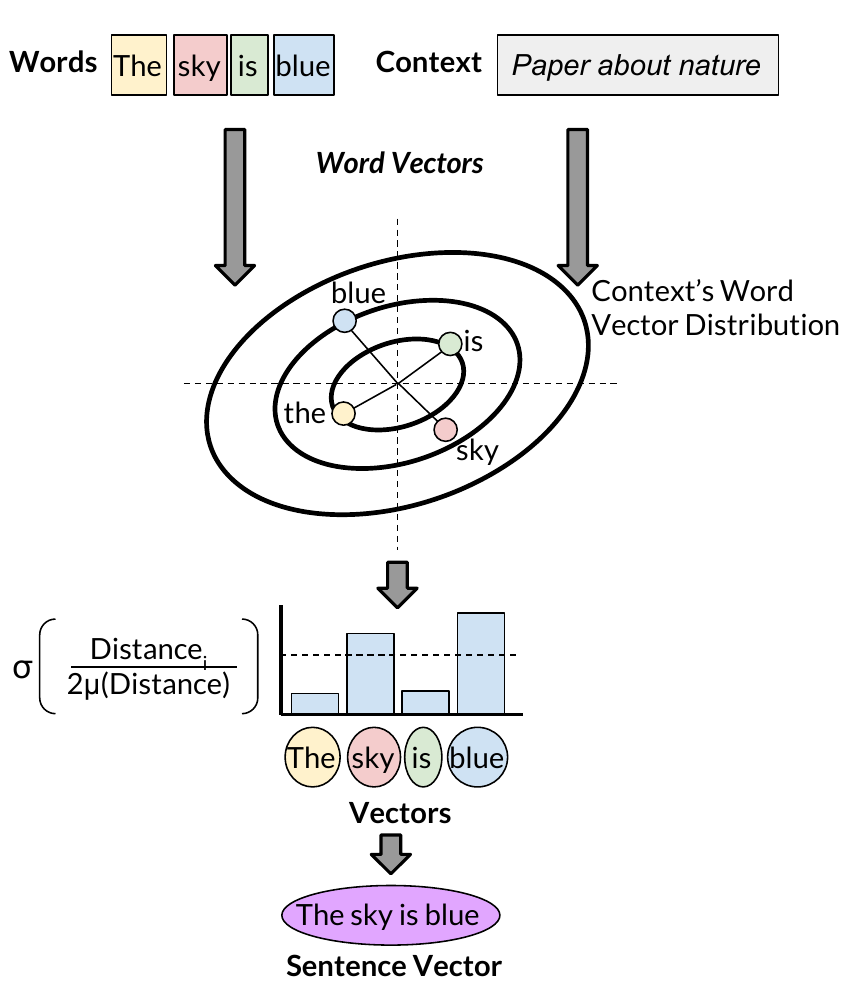}
\end{center}

\caption{Toy visualization of a sentence vector generated using contextual salience for the phrase ``The sky is blue" in the context of a paper about nature.}
\label{pull}
\end{figure}

It is generally accepted that words that are common in a document but rare in a corpus are important to the document: tf-idf is a standard technique to evaluate word importance, comparing document word frequency to overall corpus frequency, used by at least 70\% of text based recommender-systems \cite{Beel2016}. This premise is also the basis of the inverse-frequency model in Arora et al.'s ``A Simple but Tough to Beat Baseline for Sentence Embeddings," which remains nearly state-of-the-art \cite{arora2017asimple}.

Finding that CoSal correlates with tf-idf in cases where tf-idf works well (large documents and corpora), we analyze the relationship between the contextual salience of words and their contribution to a compositional representation. In order to do this, we build a vector space with both phrases and words. This space showed a clear relationship: words that are slightly more contextually salient than the others in a sentence contribute much more to the meaning of the sentence than words that are slightly less contextually salient. Combining this sigmoidal importance pattern with a weighted bag-of-words for unsupervised sentence and document representations yields better performance than widely-used supervised and unsupervised classification models on most benchmarks. Then, for especially small corpora, we extend CoSal with a method for augmenting the covariance matrix with that of a broader context.

% Since this linear combination of vectors should be in the same space as the initial vectors, one can also recursively extract words from a sentence vector. For this algorithm, we assume one of a sentence vector's neighbors is representative of the vector. Then, we develop a pathfinding algorithm from an empty word list to the word list minimizing distance to the sentence embedding.
Initially, cosine distance was used to measure the distance between a generated sentence vector and a target sentence vector, providing numerous advantages over Euclidean distance \cite{emmery_2017}. However, cosine distance still implies that all dimensions are equally valuable and uncorrelated, calculated as a dot product. Thus, it still produces context-blind measurements. Noting that M-distance can be calculated between any two points given an underlying distribution, an application of the law of cosines is used to define a robust and context-aware alternative to cosine distance, capable of realizing that ``cardinal" (A color but also a metonym for Stanford) and ``red" are less related to one another in the context of news about Stanford research than in an article about the color green. 

% \vfill
% \pagebreak
\section{Background and Related Work}
\subsection{Mahalanobis Distance} \label{mdist}
Mahalanobis distance (M-distance) is a metric to find a distance between a point and a distribution or between two points sampled from a distribution, normalized for deviation and covariance and can be interpreted as a multi-dimensional z-score \cite{10.2307/2528963}. The formula, given two points $p_1$ and $p_2$ and distribution covariance $S$ is: 

$ d(p_1, p_2, S) = \sqrt{(p_1 - p_2)^T S^{-1} (p_1 - p_2)} $ 

While M-distance has previously been applied in the context of word vectors, the uses have ranged from measuring distances between ``Gaussian word embeddings" \footnote{Word vectors as multivariate Gaussians, for spanning ambiguities. The covariance used was the joint covariance of the word embeddings} \cite{2014arXiv1412.6623V}, to a substitute for cosine distance for one-shot image recognition\footnote{This technique results in rare words being treated as distant. For example ``dog" and ``and" are both common words, and will generally have a smaller M-distance than ``Dachshund" and ``dog," even unnormalized} \cite{2017arXiv170608653R} to rule-based movie review sentiment analysis \cite{ultimatesent} or to help discriminate word sense \cite{localvector}. Consistently, the measurement is used to measure the distance between two words in a sample, not the distance of a word to the distribution. 

\subsection{tf-idf}
\begin{figure}[t!]
{\includegraphics[width=0.46\textwidth]{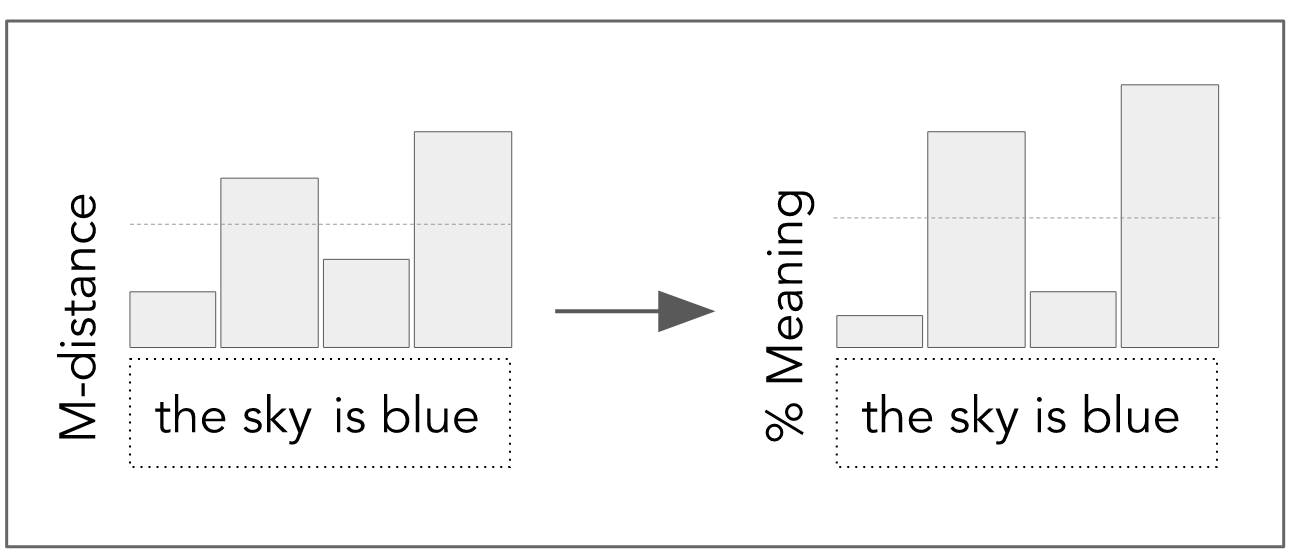}}
{\caption{A visualization of the sigmoidal relationship between the contextual salience of a word and its weight in a given sentence. \label{fig:wordProportionVis} }}
\vspace{-10px}
\end{figure}

Term frequency-inverse document frequency (tf-idf) compares a word's frequency in a document to its frequency in a corpus to determine importance. tf-idf comes with a variety of limitations. First, producing meaningful results even for common words requires a large corpus. Further, tf-idf provides no useful results for out-of-corpus tokens, limiting its usefulness in comparisons. Additionally, tf-idf is stratified for rare tokens, resulting in lines of importance as words get increasingly rare. It also ignores covariance or similarity in word meaning, and thus is highly word-choice dependent. One central idea we present in this paper is that tf-idf works because word frequency happens to be a good proxy for the unusualness of their meanings relative to their context. 

\label{performance}
\begin{table*}[t!]
\caption{\textbf{Transfer Performance.} The unsupervised state-of-the-art sentence embedding transfer learning technique, SkipThought with normalized layers is listed as ST-LN \cite{conneau2017supervised}. SkipThought, more widely used, is ST. A biLSTM was the best unsupervised model in its class (trained on unordered sentences) in \cite{conneau2017supervised}. Wang refers to Baselines and Bigrams \cite{Wang2012}, where NBSVM-bi is the best overall model, interpolating between SVM and multinomial naive Bayes with bigrams. Iyyer is the Deep Averaging Network \cite{iyyer}. The Global model uses the distance to the global average word vector while Sent uses the distance to the sentence average word vectors, single-parameter learns the sigmoid steepness for the sentence average model (per task), while ANN uses the unsupervised embeddings from the sentence average as inputs to a neural network \label{results-table}
}
\begin{center}
% \resizebox{0.9 \textwidth}{!}{%
% \begin{tabular}{lc|ccccccccc}
% \hline
% \multicolumn{2}{l}{Evaluation} & MR & MPQA & MRPC & SICK & SST & SUBJ & CR & TREC & IMDB \\ \hline
% \multirow{4}{*}{\begin{tabular}[c]{@{}l@{}}SentEval\\ (unsupervised)\end{tabular}} & tf-idf & 73.7 & 82.4 & 73.6/81.7 & - & - & 90.3 & 79.2 & 85.0 & - \\
%  & ST & 76.5 & 86.1 & 73.0/82.0 & 82.3 & 82.0 & 93.6 & 80.1 & 92.2 & - \\
%  & ST-LN & 79.4 & 89.3 & - & 79.5 & 82.9 & 93.7 & 83.1 & 88.4 & - \\
%  & BiLSTM & 77.5 & 88.7 & 73.2/81.6 & 83.4 & 80.7 & 89.6 & 81.3 & 85.8 & - \\ \hline
% \multicolumn{2}{l|}{Arora (unsupervised, ANN)} & - & - & - & 84.6 & 82.2 & - & - & - & 85 \\ \hline
% \multirow{3}{*}{\begin{tabular}[c]{@{}l@{}}Wang\\ (sup.)\end{tabular}} & NBSVM-bi & - & 86.3 & - & - & - & 93.2 & 81.8 & - & 91.2 \\
%  & SVM-bi & - & 86.7 & - & - & - & 91.7 & 80.1 & - & 89.2 \\
%  & MNB-bi & - & 86.3 & - & - & - & 93.6 & 80.0 & - & 86.6 \\ \hline
% \multicolumn{2}{l|}{Iyyer (supervised)} & - & - & - & 84.5 & 86.3 & - & - & - & 89.4 \\ \hline
% \multirow{4}{*}{CoSal (Ours)} & Global & 78.4 & 87.7 & 73.5/81.7 & 78.6 & 82.6 & 92.8 & 81.5 & 82.2 & - \\
%  & Sentence & 78.3 & 88.1 & 73.9/82.0 & 78.7 & 82.7 & 92.5 & 81.1 & 86.2 & 86.5 \\
%  & + SP & 78.4 & 88.4 & 74.1/82.3 & 78.9 & 82.8 & 92.5 & 81.2 & 86.6 & - \\
%  & + ANN & 78.4 & 88.9 & 74.1/82.3 & 80.5 & 82.8 & 92.6 & 81.2 & 89.8 & - \\ \hline
% \end{tabular}%
% }

\resizebox{0.9 \textwidth}{!}{%
\begin{tabular}{lc|ccccccccc}
\hline
\multicolumn{2}{l}{Evaluation} & MR & MPQA & MRPC & SICK & SST & SUBJ & CR & TREC \\ \hline
\multirow{4}{*}{\begin{tabular}[c]{@{}l@{}}SentEval\\ (unsupervised)\end{tabular}} & tf-idf & 73.7 & 82.4 & 73.6/81.7 & - & - & 90.3 & 79.2 & 85.0 \\
 & ST & 76.5 & 86.1 & 73.0/82.0 & 82.3 & 82.0 & 93.6 & 80.1 & 92.2 \\
 & ST-LN & 79.4 & 89.3 & - & 79.5 & 82.9 & 93.7 & 83.1 & 88.4 \\
 & BiLSTM & 77.5 & 88.7 & 73.2/81.6 & 83.4 & 80.7 & 89.6 & 81.3 & 85.8 \\ \hline
\multicolumn{2}{l|}{Arora (unsupervised, ANN)} & - & - & - & 84.6 & 82.2 & - & - & 85.0 \\ \hline
\multirow{3}{*}{\begin{tabular}[c]{@{}l@{}}Wang\\ (sup.)\end{tabular}} & NBSVM-bi & - & 86.3 & - & - & - & 93.2 & 81.8 & - \\
 & SVM-bi & - & 86.7 & - & - & - & 91.7 & 80.1 & - \\
 & MNB-bi & - & 86.3 & - & - & - & 93.6 & 80.0 & - \\ \hline
\multicolumn{2}{l|}{Iyyer (supervised)} & - & - & - & 84.5 & 86.3 & - & - & - \\ \hline
\multirow{4}{*}{CoSal (Ours)} & Global & 78.4 & 87.7 & 73.5/81.7 & 78.6 & 82.6 & 92.8 & 81.5 & 82.2 \\
 & Sentence & 78.3 & 88.1 & 73.9/82.0 & 78.7 & 82.7 & 92.5 & 81.1 & 86.2 \\
 & + SP & 78.4 & 88.4 & 74.1/82.3 & 78.9 & 82.8 & 92.5 & 81.2 & 86.6 \\
 & + ANN & 78.4 & 88.9 & 74.1/82.3 & 80.5 & 82.8 & 92.6 & 81.2 & 89.8 \\ \hline
\end{tabular}%
}

\end{center}
\end{table*}

\subsection{Context Representations}
Other research has generated context-vectors for words, calculated by an autoencoder trained on a translation model \cite{2017arXiv170800107M} or used an approach to transform word vectors based on global context using a biLSTM \cite{2017arXiv170500108P}. While techniques exist for small scale transfer learning \cite{AAAI148361}, the learning of global context \cite{2018arXiv180205365P}, and multiple-word meaning word vectors \cite{HuangEtAl2012}, this paper provides a more transparent technique which works on small contexts. 

\subsection{Arora et al.'s ``Smooth Inverse Frequency"} 
%\cite{arora2017asimple}

\label{arora}
As discussed in \cite{arora2017asimple}, a frequency-weighted average of word vectors\footnote{With this average's projection onto the corpus' sentence vectors' first principal component removed} yields a remarkably robust approximate sentence vector embedding. However, this ``smooth inverse frequency" (SIF) approach comes with practical and theoretical limitations. Practically, calculating ``estimated" word frequency using and performing common component removal are both indirect measures of contextual relevance that ignore substantial amounts of information and are thus not as ``well-suited for domain adaptation settings" as may be implied. 

Most importantly, despite the common component removal, their model still assumes that all distances in all directions in all corpora are equally significant. In results of \cite{arora2017asimple}, the lower-than-expected performance on sentiment analysis is attributed to the issue of the distributional hypothesis: words related to sentiment show up often together so are close as vectors. A variation to our sentence model draws from \cite{arora2017asimple}, incorporating the distance to the sentence's average instead of the context average, using the context's covariance. \cite{arora2017asimple}'s calculation of word frequencies from the unigram count of words\footnote{Which they note is calculated using various corpora of online text including Wikipedia and political blogs} also means that their approach does not work for out-of-vocab words and can't be generated from the word vectors alone, suffering some of the issues of tf-idf.

\section{Method}
In short, contextual salience is the distance between two points in a word vector space when accounting for the correlation and variance of the dimensions of the word's context. For sentence representations, every word's weight is the sigmoid of its contextual salience, relative to the mean salience of the words in the sentence, though the salience measure itself is more widely applicable. This is visualized in Figure~\ref{pull}.

While global context is powerful, that approach alone is limited when dealing with small contexts: We hope to place less emphasis on features like tense or plurality that may be consistent in a text but vary in language or a larger contextual corpus. 

For this, mirroring tf-idf, we weight the covariance of the document and the covariance of a larger corpus, roughly corresponding to tf and idf respectively. Then, we take the (element-wise) geometric mean of the weighted covariance matrices, as the document distribution is theoretically drawn from a specific subset of the corpus distribution, and this transformation emphasizes the relationships that are significant in both the corpus distribution and in the weighted document distribution. Letting S' be the weighted covariance S, CoSal is calculated as follows:
$$CoSal(v) = \sqrt{v^T (S'_{doc} \odot S'_{corp})^{-1} v}$$ 
Like tf-idf, CoSal can be used with arbitrary weighting schemes; that is, the interpolation applied between the document and corpus can mirror most tf-idf weighting schemes. The recommended weighting scheme, draws from the common ``augmented" \cite{hinrich_2009} tf-idf weighting scheme\footnote{The ``natural" \cite{hinrich_2009} scheme, which is essentially just tf, is used in this paper for very large corpora},  with $\sqrt[^+]{x}$ as the sign-preserving square root $$S'_{corp} = \sqrt[^+]{S_{corp}} \mbox{ and } S'_{doc} = \sqrt{\frac{|S_{doc}| + |S_{corp}|}{2}}$$

\begin{algorithm}[b!]
   \caption{Training: Finds the distribution of the document's word vector cloud}
   \label{alg:example1}
\begin{algorithmic}
{
 \fontsize{9}{6}
 \STATE {\bfseries Input:} $trainingSet$, a large document or transfer learning training set
\STATE {\bfseries Result:} Returns Mahalanobis metric of training set and average of training word vectors
\STATE $vecs = $[$getVec(w_i)$ {\bfseries for} word $w_i$ {\bfseries in} $trainingSet$]\;
\STATE   $globalAvg$ = avg($vecs$)\;
\STATE   $metric$ = MahalanobisMetric(cov($globalCovariance$))\;
\STATE {\bfseries Return:} $globalAvg$, $metric$
%  \caption{. }
}
\end{algorithmic}
\end{algorithm}

\subsection*{Confidence}
We can also generalize $S'_{doc} = \sqrt{p|S_{doc}| + (1-p)|S_{corp}|}$, where $p$ corresponds to the representativeness of the smaller context, logarithmically proportional to the number of words in the context. 

\subsection*{Sentence Average}
Using the sentence average in place of the global average improves performance because it makes a useful approximation: the more important the difference is between the word and the approximate meaning of the sentence, the more role that word likely plays in defining the sentence. This helps account for shortcomings of the distributional hypothesis (Discussed in~\ref{arora}), since dimensional covariances which are typically smaller but prominent in a particular corpus (For example, sentiment-related words in a sentiment corpus) will be automatically identified as important.

\section{Performance}

\subsection{Evaluation}
The benchmarks perfomed include subjectivity detection (SUBJ), product reviews (CR), entailment (SICK), opinion polarity (MPQA), question-type identification (TREC), movie reviews (MR), paraphrase detection (MRPC), semantic similarity (STS) through Facebook's SentEval \cite{conneau2017supervised} suite of tests as well as the IMDB review dataset. To evaluate the quality of the transfer-learned sentence embeddings, the benchmarks perform a linear regression classification on the returned sentence vectors. Unlike in previous sections, fastText\footnote{2m vocabulary 300d word vectors trained w/ Common Crawl} \cite{mikolov2018advances} is used, which has less predictive word vectors \cite{conneau2017supervised}, but supports character-level word vector prediction for misspelled or esoteric words\footnote{\citet{ajay_patel_2018_1196590} was used for fast vector lookup}. The downstream tests were performed by feeding the unsupervised sentence embeddings to a neural network for supervised classification. Because the word vector model contained fewer capitalized words, lowercased performance was generally better than case-sensitive performance.

\subsection{Procedure}
The following Algorithms~\ref{alg:example1}~and~\ref{alg:example2} include the sentence variation discussed earlier in this paper as well as the global context only version, which measures to the global average instead of the sentence average. The sentence variation approximates the sigmoid linearly, as the distances in the sentence case tend to cluster more. 

% \subsection{Results}

\begin{algorithm}[b!]
 \caption{Sentence embedding algorithm. Returns the sentence vector in a trained context}
  \label{alg:example2}
\begin{algorithmic}
{
 \fontsize{9}{6}
\STATE {\bfseries Input:} $sentence$: a list of words; $globalAvg$ and $metric$ from training algorithm; $useGlobal$: a boolean for whether to compare to the global average or the sentence's average word vector.
\STATE {\bfseries Result:} Sentence embedding, weighting words by their contextual importances 
\STATE    $vecs =$ [$getVec(w_i)$ {\bfseries for} word $w_i$ {\bfseries in} $sentence$] \;
\STATE    $avgVec = globalAvg$ {\bfseries if} $useGlobal$ {\bfseries else} unit(avg($vecs$)) \;
\STATE    $dists = [metric$.dist($vec$, $avgVec$) {\bfseries for} $vec$ {\bfseries in} $vecs$]\;
\STATE    $dists$ /=$ 2 *$ avg($dists$) \;
\STATE    $weights =$ sigmoid$(dists)$\;
\STATE {\bfseries Return:} sum($weights_i$ * $vecs_i$ {\bfseries for} $i=0$ {\bfseries to} len(vecs))\;
}
 \end{algorithmic}
\end{algorithm}

\begin{figure*}
\begin{center}
\includegraphics[width=0.43\textwidth]{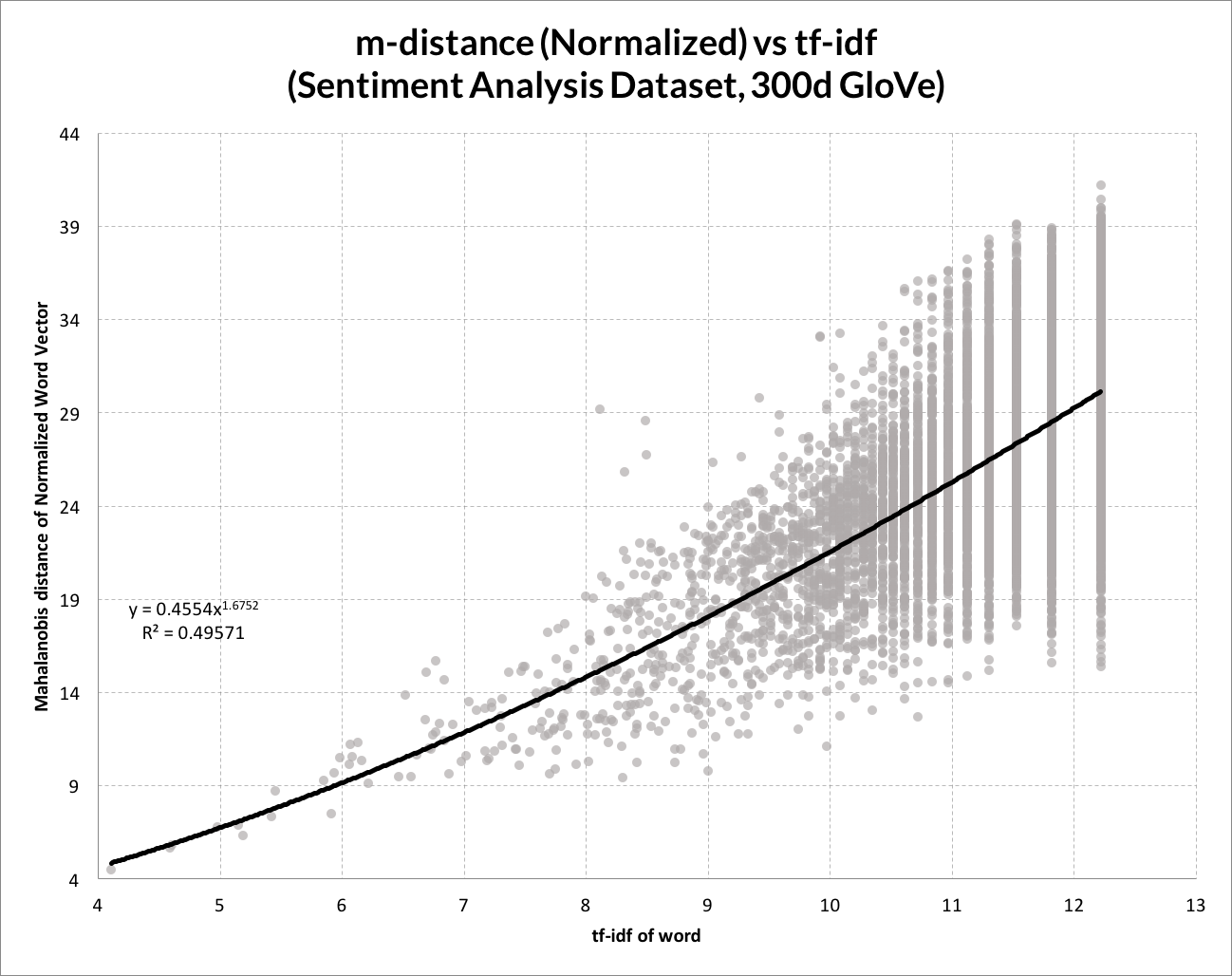} 
\includegraphics[width=0.43\textwidth]{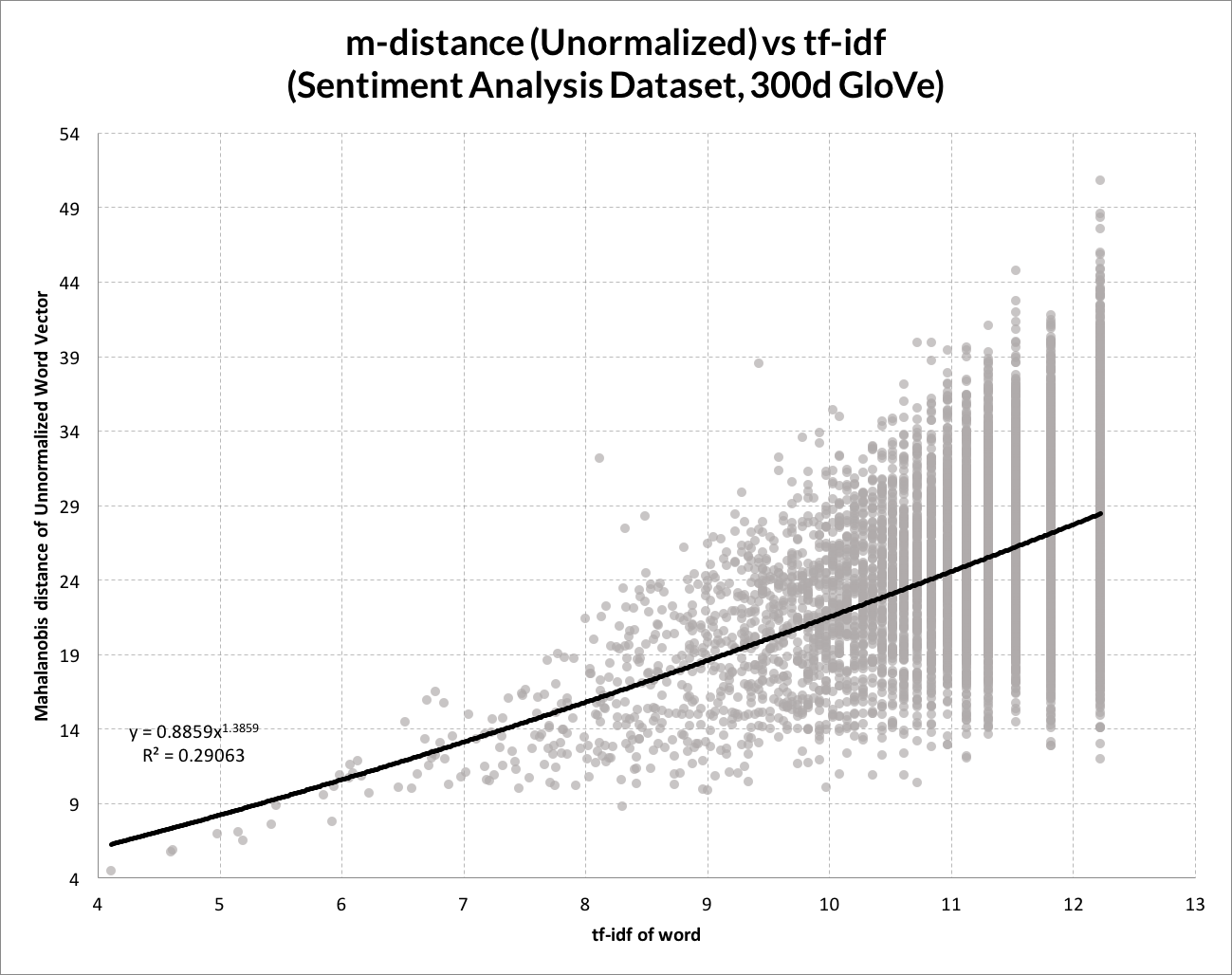}
\end{center}

\caption{\textbf{Relationship of Mahalanobis distance to tf-idf} on the Stanford Treebank Dataset as a large document with 300d GloVe word vectors, normalized (left) and unnormalized (right) \label{tfidf}}
\end{figure*}

\subsection{Analysis}
Notably, while this model performs comparably or better than SkipThought with Layers Normalized (ST-LN) in almost all tests, and outperforms the more widely used SkipThought on most \cite{conneau2017supervised}, it is vastly more efficient with computationally and data efficient. It should be some cause for concern 

\subsubsection{Efficiency}
This model requires orders of magnitude fewer sentence examples than models like SkipThought or the bidirectional LSTM to produce high quality results and no training beyond that of the underlying word vectors. One question is how many samples are needed to establish a context. This model's performance exceeds tf-idf's maximum performance for most tests after only 10 sentences. Further, every example seems to improve performance logarithmically. On the MRPC dataset, 99\% of the model's ultimate accuracy is reached with 60 sentences, with 5, 10 and 30 example sentences resulting in accuracies of 94\%, 96\% and 97\% of the ultimate performance. 

On a 2015 Macbook Pro without a GPU, it takes 22ms per thousand sentences in the training dataset to ``train" the model given unique training dataset word vectors\footnote{That is, to calculate their dimensional covariance}, requiring less than 200ms for all datasets listed besides IMBD and SST. On the other hand, on the models for which training curves are available, where Skipthoughts-LN outperforms CoSal-sentence model, it took, on average, approximately a week of training on more powerful hardware for them to earn comparable scores\cite{2016arXiv160706450L}. The ST-LN model used for comparison in the table was trained for a month. 
\subsubsection{Stability}
Finally, notably, while the single-parameter (SP) adjusted sigmoids had marginally better test performance in most cases, the size of the difference tended to be small (an average improvement of $< 0.2\%$). This indicates that this baseline is robust.

\section{Derivation}
% \subsection{Empirical Derivation}
First, the Mahalanobis distance of words in a corpus was compared to their tf-idf. The context used here was the Stanford Sentiment Analysis Treebank dataset \cite{Socher2013RecursiveDM}, treated as a document for the sake of a maximal comparison. Both normalized and unnormalized GloVe word vectors\footnote{300d vectors from 42B token Common Crawl} \cite{pennington2014glove} were compared, as shown in Figure~\ref{tfidf}. The normalized approach resulted in substantially better correlation ($r^2 = 0.496$ vs $0.291$) between a word's tf-idf and Mahalanobis distance, presumably because the length of the original word vectors was correlated with word frequency, skewing the vector distribution's density. 

We modified the GloVe algorithm, a widely used approach to generate word-vectors \cite{simon_2017}, to create a single vector space which contained vectors corresponding to both two-word phrases and words. We used spaCy \cite{honnibal-johnson:2015:EMNLP} to identify two-word grammatical phrases, randomly treating a two-word phrase as a single token 50\% of the time, and the remaining times replacing it with one of the underlying words\footnote{To prevent the phrase's sentence position from biasing its word vector, given the reduced context at the beginning or end of a sentence}. We then calculated the linear combination of the vectors of the two words composing a phrase which was closest to the phrase vector. Then the relationship between the proportion of the first word in the linear combination and the proportion of the first word of the Mahalanobis distance was plotted to produce the curve in Figure~\ref{fig:wordProportion} in the appendix.

To extend the two-word pattern to a sentence embedding model, the word vector's distances are rescaled so that their mean distance corresponds to the sigmoid's center, as visualized in Figure~\ref{fig:wordProportionVis}. 

% \vfill

% \subsection{Statistical Derivation}
% Suppose words, phrase, and sentence vectors exist in the same vector space with a similar D-dimensional probability distribution $W,H,S \sim A \in R^D$, respectively. We assume that words $w_1, w_2$ to express a two-word phrase, $h$, are chosen by selecting one of the closest word vectors $v_1$ to its phrase vector $v_h$, with the remainder completed by the second word vector $v_2$. 

% The expected distance to the nearest word to a sentence vector would be, with a smooth probability distribution and a large vocabulary $V$, approximately the $D-th$ root of the inverse density of words around the point: $l = \sqrt[D]{\frac{1}{|V| p(v_h)}}$. We aim to find the expected proportion of the meaning in a phrase attributed to a word as a function of the proportion of surprisal (corresponding to the negative log likelihood) of the word, closely correlated to that of the phrase. If we approximate the proportion of meaning of the second word vector $\frac{d(v_2, v_h)}{d(v_1, v_h) + d(v_2, v_h)} \approx \mathbf{E}\left[\frac{d(v_2, v_h)}{d(A_\mu, v_h) + d(v_2, v_h)}\right]$, and the proportion of surprisal is 

\section{Conclusion} 
Contextual salience provides a fast and transparent mechanism to account for the variation in the importance of features of a text in differing contexts. While the benchmarks and methods presented in this paper demonstrate its usefulness, we believe CoSal has broad applications and can be easily integrated into a variety of natural language algorithms. For example, the base contextual salience technique can likely be augmented with contextual word vectors \cite{2018arXiv180205365P} and could be used as an additional input to a recurrent neural network to account for order. One extension would be to combine this approach with ELMo vectors, which generate word vectors for words based on the contexts in which they appear \cite{Gardner2017AllenNLP}.

% Additionally, the recursive document embedding pattern in Subsection ~\ref{decomp} suggests consolidation of information can aid abstraction, and our proposed metric allows for the contextual salience of sentence vectors to be measured where no simple frequency count would work.

Beyond accurate sentence embeddings to augment language-processing algorithms, a technique to extract meaning from generated vectors, and new metrics, the results in this paper may carry implications about the compositional structure of language. The idea that a representation of context allows one to consistently disregard and overemphasize information when analyzing collective meanings can be broadly applied. It suggests that, although all parts play a role in establishing collective meaning of several related pieces of information, the presence of a few salient details plays a disproportionate role. 

\renewcommand\refname{References}
\bibliography{references} 
\bibliographystyle{icml2019}
	
\appendix
\onecolumn
\vfill
\pagebreak
\section*{Appendix}
\section{Context-Adjusted Cosine Distance}
Due to the context-blind nature of cosine distance, we also suggest an approach to account for the underlying vector distribution. Treating the CoSal of the difference between the words as the opposite leg of the triangle and treating the word's salience measures as the legs, the law of cosines is solved for the cosine of the angle between the words. Where a and b are the legs (the distance between the word and the average vector) and c is the opposite side (the distance between the two words), we calculate $cosC = \frac{a^2 + b^2 - c^2}{2ab}$.  

This approach corresponds to the following intuition: The more salient the difference between the meanings of two words is relative to their individual saliences $(c/ab)$, the farther apart the meaning of the two words are. For example, in the context of an article about a development in Stanford self-driving car tech, ``cardinal" and ``red" have a distance of 0.943, while in random excerpts from a Wikipedia article about the color green \cite{wikipedia_2018}, ``cardinal" and ``red"  have a distance of 0.909.

% \begin{figure}[h!]
% \begin{center}
% \includegraphics[width=0.4\textwidth]{pathfinding.png} 
% \end{center}
% \caption{A pathfinding visualization. Each next word is one of those closest to the remaining vector.}
% \end{figure}

\section{Short Text Visualization}
\begin{figure*}[h!]
\begin{center}
\includegraphics[width=0.95\textwidth]{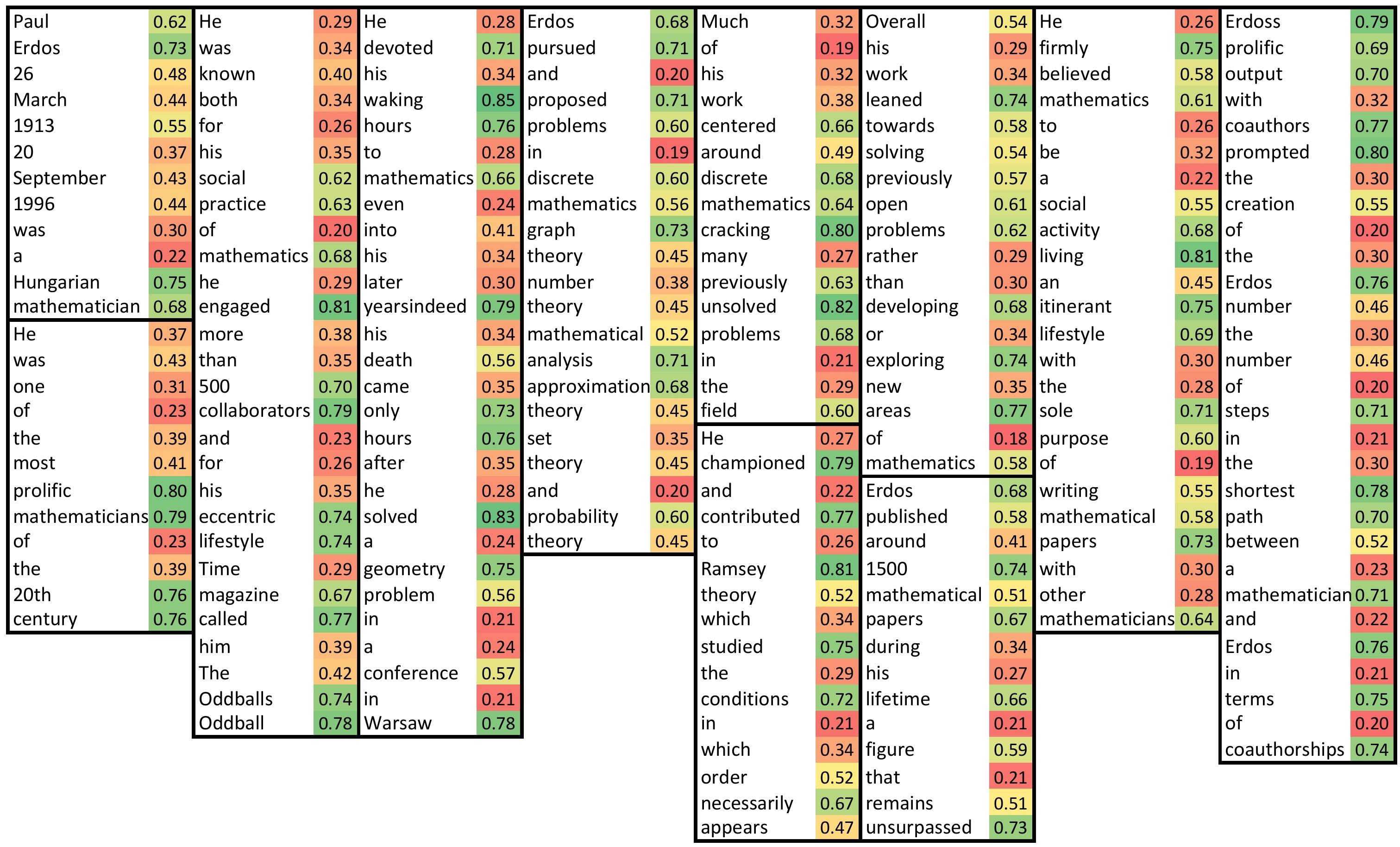} 
\end{center}
\caption{The global sigmoid weights generated for a short excerpt about Erdos from Wikipedia \cite{wikipaul}, using only the shown text as global context and TREC for corpus context, using the recommended weighting with p = 0.2}
\label{erdosfig}
\end{figure*}

We include an example of the CoSal weights being applied to a set of sentences, when given a short context about Erdos in Figure~\ref{erdosfig}. 

\vfill 
\pagebreak

\section{Hyperparameters}
\begin{figure}[h]
{\caption{\textbf{The proportion of a two-word phrase's vector that is determined by a given word, as a function of the word's proportion of the sum of the two words' importances.}, 
Each two-word phrase was identified by spaCy \cite{honnibal-johnson:2015:EMNLP} from the Stanford Sentiment Analysis Treebank \cite{Socher2013RecursiveDM}.
The proportions are shown with a moving average (gray) and a sigmoid (black, erf) fit to the data. The y-axis is calculated by finding the x such that $x(v_1) + (1-x)(v_2)$ is as close to the phrase vector as possible. The x-axis is 
$\frac{ d(v_1, v_{avg}) }{ d(v_1, v_{avg}) + d(v_2, v_{avg}) }$
}\label{fig:wordProportion}}
{
\begin{center} \includegraphics[width=0.5\textwidth]{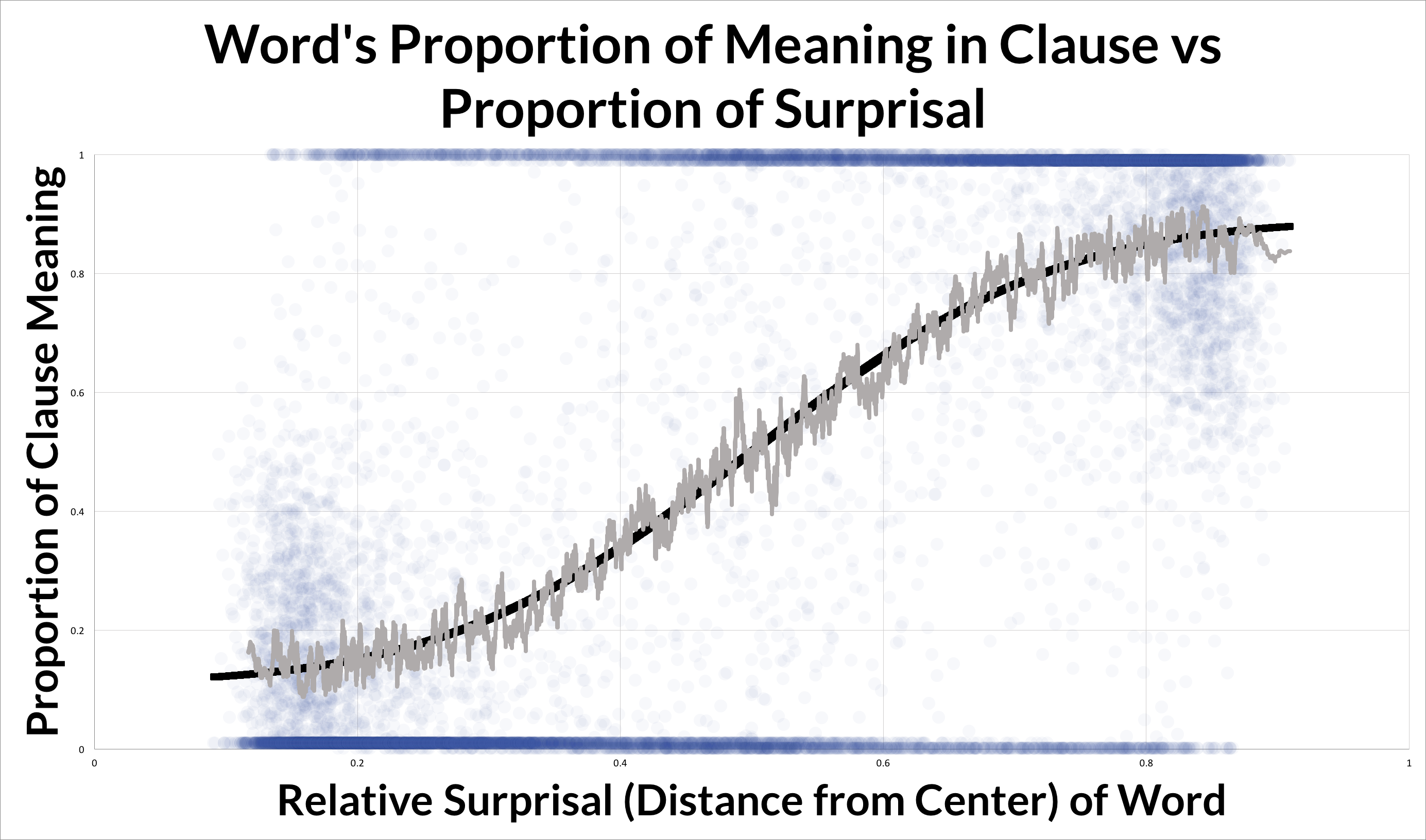}
\end{center}
}
\end{figure}

The sigmoid used throughout the paper is motivated by and derived from that in Figure~\ref{fig:wordProportion}, scaled vertically by 0.7, horizontally by 0.11 (slope = 1.58), and the resulting sentence vector is normalized. The single-parameter (SP) sigmoids have the same vertical scaling, but have steepnesses optimized per-task (CR = 0.11, MR = 0.12, SST = 0.14, SICK = 0.17, MRPC = 0.18, SUBJ = 0.22, MPQA = 0.24, TREC = 0.26). The neural network used in the ANN results had two hidden layers with 150 neurons, ReLU activation functions, a dropout probability of 0.3, and layer normalization. As shown in Table 1, the sensitivity to tuning is low.

Note that the shape of the curve is almost identical when the phrases containing stop words are removed. Stop words can be approximately calculated as the least important 15\% of words by distance.

\end{document}